\documentclass{article}

\usepackage[preprint]{neurips_2025}

\usepackage[utf8]{inputenc} 
\usepackage[T1]{fontenc}    
\usepackage{hyperref}       
\usepackage{url}            
\usepackage{booktabs}       
\usepackage{amsfonts}       
\usepackage{nicefrac}       
\usepackage{microtype}      
\usepackage{xcolor}         
\usepackage[table]{xcolor}  
\usepackage{colortbl}    
\usepackage{multirow}
\usepackage{amsmath}
\usepackage{graphicx}
\usepackage{bm}
\usepackage{wrapfig}
\usepackage{algorithm}
\usepackage{algorithmic}
\usepackage{enumitem}
\usepackage{wrapfig}
\usepackage{amssymb}
\usepackage{pifont} 
\usepackage{amsthm}
\newtheorem{proposition}{Proposition}[section]

\newtheorem{definition}{Definition}[section]

\definecolor{mygreen}{RGB}{56, 87, 35}
\definecolor{myred}{RGB}{142, 0, 0}

\newif\ifhideproofs
\ifhideproofs
  \usepackage{environ}
  \NewEnviron{hide}{}

\fi

\title{Towards Pre-trained Graph Condensation \\via Optimal Transport}

\author{%
  Yeyu Yan$^1$,~ 
  Shuai Zheng$^1$,~
  Wenjun Hui$^1$,~
  Xiangkai Zhu$^2$,~
  Dong Chen$^1$,~\\
 \textbf{Zhenfeng Zhu}$^1$\thanks{Corresponding author}~~,
  \textbf{Yao Zhao}$^1$,~
  \textbf{Kunlun He}$^3$~
  \\
  $^1$Beijing Jiaotong University, China\\
  $^2$Jinan University, China~~
  $^3$Chinese PLA General Hospital, China\\
  \texttt{yanyeyu-work@foxmail.com, zs1997@bjtu.edu.cn,}\\
  \texttt{22110090@bjtu.edu.cn, 18063597830@163.com, dchen2001@bjtu.edu.cn, }  \\
  \texttt{zhfzhu@bjtu.edu.cn, yzhao@bjtu.edu.cn, kunlunhe@plagh.org}  \\
  }

\begin{document}

\maketitle

\begin{abstract}

%
%
Graph condensation (GC) aims to distill the original graph into a small-scale graph, mitigating redundancy and accelerating GNN training. 
However, conventional GC approaches heavily rely on rigid GNNs and task-specific supervision. Such a dependency severely restricts their reusability and generalization across various tasks and architectures.
%
%
In this work, we revisit the goal of ideal GC from the perspective of GNN optimization consistency, and then a generalized GC optimization objective is derived, by which those traditional GC methods can be viewed nicely as special cases of this optimization paradigm.
Based on this, \textbf{Pre}-trained \textbf{G}raph \textbf{C}ondensation (\textbf{PreGC}) via optimal transport is proposed to transcend the limitations of task- and architecture-dependent GC methods. Specifically, a hybrid-interval graph diffusion augmentation is presented to suppress the weak generalization ability of the condensed graph on particular architectures by enhancing the uncertainty of node states. 
Meanwhile, the matching between optimal graph transport plan and representation transport plan is tactfully established to maintain semantic consistencies across source graph and condensed graph spaces, thereby freeing graph condensation from task dependencies.
%
To further facilitate the adaptation of condensed graphs to various downstream tasks, a traceable semantic harmonizer from source nodes to condensed nodes is proposed to bridge semantic associations through the optimized representation transport plan in pre-training.
%
Extensive experiments verify the superiority and versatility of PreGC, demonstrating its task-independent nature and seamless compatibility with arbitrary GNNs.

\end{abstract}

\section{Introduction}

Graph neural networks (GNNs) have emerged as a powerful approach to graph data analysis, showing excellent efficacy in a variety of domains, such as recommender systems \cite{DBLP:conf/www/ChenBSXZHHWH24}, text analysis \cite{DBLP:conf/aaai/Yin024, DBLP:journals/tbd/YanZYYL24}, social network analysis \cite{DBLP:journals/eswa/YanLYLZ23}, and so on.
However, the dual challenges of exponentially growing graph data volumes \cite{DBLP:conf/kdd/ZhangYS0OLT0022} and increasingly sophisticated GNN architectures \cite{DBLP:journals/pami/ZhengZLLZ24} present significant obstacles for GNN training. These challenges become particularly acute in scenarios that require multiple GNN training, such as continual graph learning \cite{DBLP:conf/nips/HanF024} and federated graph learning \cite{openfgl_arxiv24}.

\begin{wrapfigure}{r}{8cm}
\vspace{-0.em}
    \centering
    \includegraphics[width=0.56\textwidth]{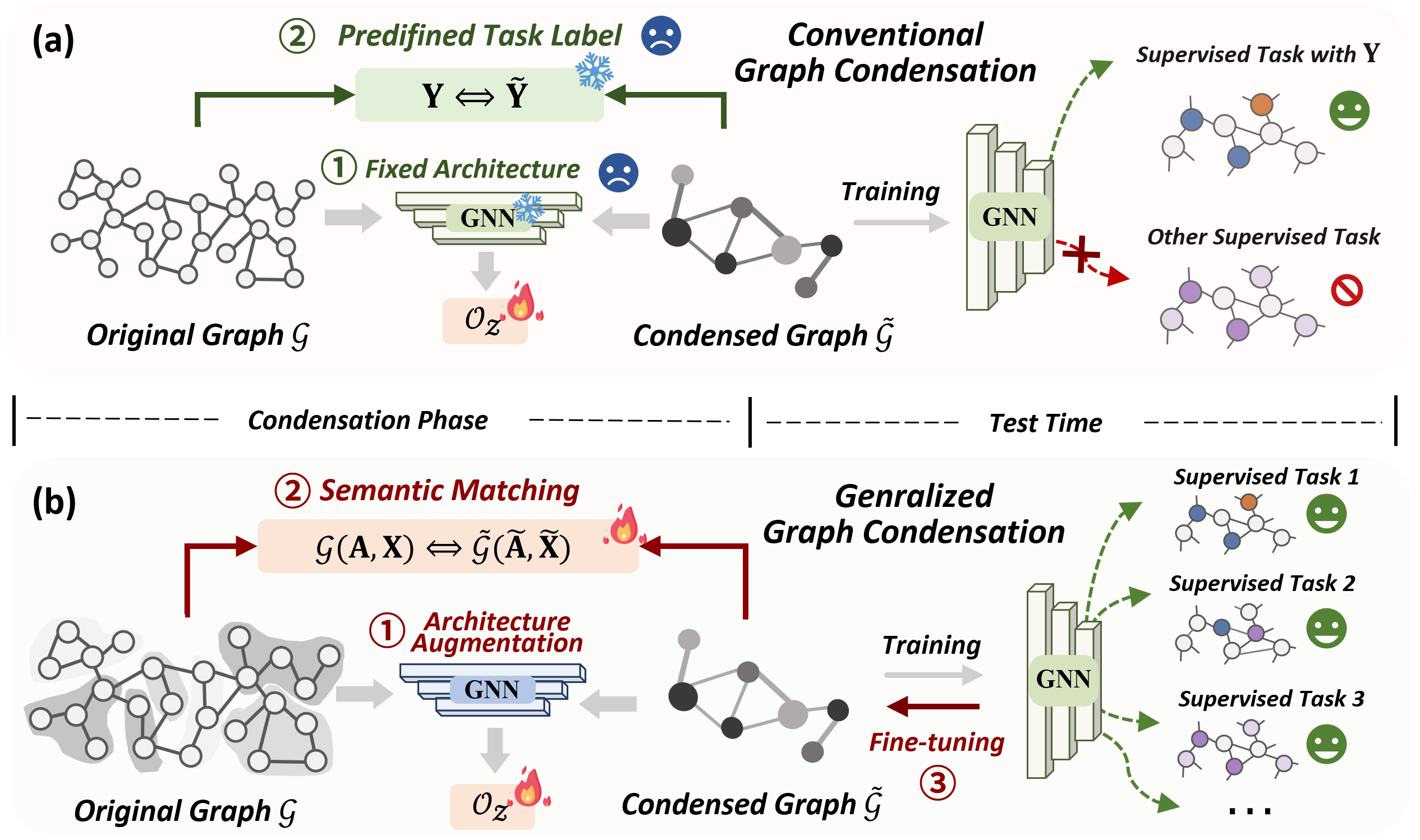}
    \vspace{-0.em}
\caption{(a) The conventional GC framework. Due to specific \textcolor{mygreen}{\ding{172}} architecture and \textcolor{mygreen}{\ding{173}} task dependencies, $\tilde{\mathcal{G}}$ lacks generalization and reusability. (b) Our proposed generalized GC framework. It solves the above problems by \textcolor{myred}{\ding{172}} and \textcolor{myred}{\ding{173}}, and  fine-tuning $\tilde{\mathcal{G}}$ in specific scenarios by \textcolor{myred}{\ding{174}}.}
    \label{fig:pipeline}
    \vspace{-1em}
\end{wrapfigure}

In response to the above issues, graph condensation (GC) \cite{gcond_iclr22, sfgc_nips23} is proposed to optimize and synthesize a small-scale condensed graph from the real graph. 
The core objective is to ensure that condensed graphs achieve comparable performance as original graphs when training the same GNN. This approach significantly reduces computational overhead and becomes an up-and-coming solution for efficient graph learning. 
Depending on matching strategies, GC fall mainly into three types.
For example, \cite{gcond_iclr22, doscond_kdd22, sgdd_nips23} distilled the properties of the real graph by mimicking its gradient changes in GNN training. \cite{sfgc_nips23, geon_icml24} replicated the training trajectory on original graphs to synthesize condensed graphs. Other studies \cite{gcdm_arxiv22, cgc_www25} generated the graph by aligning the latent space representations of both graphs and maintaining the distribution consistency. 
Recently, \cite{cgc_www25} pointed out that all these methods can be generalized as distribution matching, with the basic pipeline shown in Fig. \ref{fig:pipeline} (a).

Despite these significant efforts, current methods suffer from two critical limitations: \textbf{(i) Architecture-bound optimization.} Current methods can be regarded as condensing through constraints in the same representation space \cite{cgc_www25}. Therefore, they have to adopt a fixed encoder to map nodes from the graph space to the representation space in condensation.
Such an excessive dependence on a specific architecture can degrade the generalization of condensed graphs, leading to poor performance on other models.
In addition, the learnable model parameters inherit task bias \cite{selfgc_kdd24}, which causes another challenge.
\textbf{(ii) Task-constrained condensation.} To simplify optimization \cite{gcond_iclr22}, existing methods typically assume that tasks and labels are known, thus the effectiveness of GC depends heavily on the guidance of supervision signals. Unfortunately, this assumption often proves impractical in real-world scenarios. 
For example, in social networks, multiple tasks may coexist (e.g., preference classification, income prediction, and relation prediction) \cite{DBLP:conf/icml/FeyHHLR0YYL24}, and user labels frequently vary owing to privacy constraints and annotation discrepancies \cite{openfgl_arxiv24}. %
When the task or label changes, conventional GC methods require re-condensation to capture new knowledge, significantly diminishing the reusability of condensed graphs.
Although \cite{seflgc_arxiv24} attempted to address label dependency by contrastive learning, its reliance on class similarity hinders its application to other supervised tasks. 

To tackle the two critical challenges altogether, we first revisit the GC objective based on the goal of condensed graphs. It aims to achieve comparable performance to the original graph when training the same GNN (i.e. GNN optimization consistency). From this perspective, we derive a generalized GC optimization paradigm without task constraints as shown in Fig. \ref{fig:pipeline} (b) and design a flexible and generalized GC method termed \textbf{Pre}-trained \textbf{G}raph \textbf{C}ondensation (PreGC).  
Inspired by data augmentation \cite{DBLP:conf/icde/CuiCYDF024}, graph diffusion augmentation is proposed to perturb node diffusion states via hybrid intervals, thereby enhancing the generalizability of condensed graphs and mitigating their dependence on specific architectures.
To decouple the reliance on task labels, we present a novel matching mechanism, named transport plan matching.
It maintains consistency in semantic associations between graph space and representation space, thus guaranteeing a unique mapping from each source node to its condensed counterpart.
More importantly, transport plans provide explicit feedback on the significance of the source node, alleviating the poor traceability and interpretability issues in conventional condensation methods.
After pre-training, a traceable semantic harmonizer via the optimal transport plan is derived, 
enabling flexible transfer of task signals from the original graph to the condensed graph for adaptation to diverse tasks.

\textbf{To the best of our knowledge}, PreGC is the first work for generalized graph condensation considering  both architecture- and task-agnostic scenarios. Our main contributions are outlined as follows:
\begin{itemize}[itemsep=2pt,leftmargin=*,topsep=2pt]

\item Formal theoretical analysis reveals that conventional GC methods are inherently governed by the reconstruction term and the fitting term.
Inspired by such an insight, we propose a pre-training framework of graph condensation, generalizing the condensed graph to more scenarios.


\item 
To empower the condensed graph for architecture-agnostic scenarios, a hybrid-interval graph diffusion augmentation is presented, which enhances the diversity of node representations by introducing stochasticity into the diffusion process of both the condensed and original graphs.

\item 
We propose an optimal transport plan matching to realize semantic alignment without task constrained. By keeping a consistent optimal plan for graph alignment and representation alignment, 
the uniqueness of semantic associations between source nodes and condensed nodes is guaranteed.


\item Experiments on four mainstream tasks and nine representative GNN architectures demonstrate the superiority and generalization of PreGC. 
Further data valuation of the original graph highlights the PreGC's excellent traceability and interpretability.


\end{itemize}

\vspace{-0.2em}
\section{Preliminary}
\vspace{-0.2em}

\textbf{Graph Condensation (GC).} 
Given a large-scale original graph $\mathcal{G}= (\mathbf{A},\mathbf{X})$ with a node set $\mathcal{V}$ ($|\mathcal{V}|=N$), where $\mathbf{A} \in \mathbb{R}^{N\times N}$ denotes the symmetrically normalized adjacency matrix and $\mathbf{X} \in \mathbb{R}^{N\times f}$ is a $f$-dimensional node feature matrix. GC \cite{gcond_iclr22, gdem_icml24} aims to condense a small synthetic graph $\tilde{\mathcal{G}}= (\tilde{\mathbf{A}},\tilde{\mathbf{X}})$ with a condensed node set $\tilde{\mathcal{V}}$ ($|\tilde{\mathcal{V}}|=M$) form the real large graph $\mathcal{G}$, where $\tilde{\mathbf{A}} \in \mathbb{R}^{M\times M}$, $\tilde{\mathbf{X}} \in \mathbb{R}^{M\times f}$, and $M \ll N$. At the same time, trained GNNs with the specific task have comparable performance on both $\tilde{\mathcal{G}}$ and $\mathcal{G}$, thus accelerating the training of GNNs.

\textbf{Graph Optimal Transport.}
Optimal transport (OT) \cite{WD_access25} aims to seek the most cost-efficient transport plan $\pi$ that transforms a source distribution $\mu$ into a target distribution $\nu$ while minimizing the total transport cost (i.e., the optimal transport distance). The elements of $\pi$ represent the probability of mass transferring from one location to another. In this work, we primarily concentrate on graph optimal transport \cite{DBLP:conf/icml/ChenG0LC020, FGWD_algorithms20}, which extends the OT to compare structured data. 
Given two graphs $\mathcal{G}=({\mathbf{A}},\mathbf{X})$ and $\tilde{\mathcal{G}}= (\tilde{\mathbf{A}},\tilde{\mathbf{X}})$ with the nodes' empirical distribution $\mu \in \mathbb{R}^N$ and $\nu \in \mathbb{R}^M$, the general form of the graph optimal transport distance can be formalized as:
\begin{equation}\label{eq_g_opt}
\mathcal{W}(\mathcal{G},\tilde{\mathcal{G}})=\mathop{\rm min} \limits_{\pi\in{\Pi}(\mu,\nu)} \big\langle \mathcal{C}(\mathbf{X},\tilde{\mathbf{X}},\mathbf{A},\tilde{\mathbf{A}},\pi),\pi  \big\rangle
\vspace{-0.2em}
\end{equation}
%
where $\Pi(\mu, \nu) = \{\pi \in \mathbb{R}^{N\times M}| \pi \mathbf{1}_M = \mu,\mathbf{1}_N\pi  = \nu \}$ is set of the joint distributions $\pi$ with marginals $\mu$ and $\nu$, and and $\langle \cdot, \cdot \rangle$ denotes the inner product for matrices. $\mathcal{C}$ represents the cost function that quantifies the cost of transporting mass from elements in one domain to another. 


The transport distance quantifies the distribution divergence between condensed and original graphs, offering a novel optimization approach for GC. Meanwhile, the transport plan establishes explicit node correspondences, overcoming the limitations of poor traceability in conventional GC methods.

\vspace{-0.2em}
\section{Revisiting and Generalizing Graph Condensation}\label{revisiting}

In this section, we revisit the objective of graph condensation and reveal the limitations of existing GC methods, thus motivating the design of PreGC.


\textbf{Revisiting.} For convenience, we follow \cite{gdem_icml24} and start with a vanilla example, which adopts a $K$-order SGC \cite{sgc_icml19} as the GNN and simplifies the objective of GNNs into the MSE loss:
$\mathcal{L}_{MSE}=\|\mathbf{A}^K\mathbf{X}\mathbf{W}-\mathbf{Y}\|_F^2$, where $\mathbf{Y} \in \mathbb{R}^{N\times C}$ denotes the target variable and $\mathbf{W} \in \mathbb{R}^{f\times C}$ is the model parameter. For the original graph $\mathcal{G}$, the optimal $\mathbf{W} = \mathop{\rm arg~min} \limits_{\mathbf{W}}\|\mathbf{A}^K\mathbf{X}\mathbf{W}-\mathbf{Y}\|_F^2$ can be obtained by MSE loss. 
In general, the goal of condensation is that the GNN trained on the condensed graph $\tilde{\mathcal{G}}$ to achieve comparable performance to those trained on $\mathcal{G}$, i.e
\begin{equation}\label{eq_dis_z}
\min \|\mathbf{W}- \tilde{\mathbf{W}}\|_F=\min \|(\mathbf{A}^K\mathbf{X})^\dagger \mathbf{Y}- (\tilde{\mathbf{A}}^K\tilde{\mathbf{X}})^\dagger \tilde{\mathbf{Y}}\|_F
\end{equation}
where $\tilde{\mathbf{W}}= \mathop{\rm arg~min} \limits_{\tilde{\mathbf{W}}}\|\tilde{\mathbf{A}}^K\tilde{\mathbf{X}}\tilde{\mathbf{W}}-\tilde{\mathbf{Y}}\|_F^2$ with target variable $\tilde{\mathbf{Y}} \in \mathbb{R}^{M\times C}$ on $\tilde{\mathcal{G}}$, and $\dagger$ denotes pseudo inverse. Subsequently, we further derive the following proposition (Proofs are in Appendix A):
\begin{proposition}\label{proposition_1}
Suppose that it has an analytical filter $g(\cdot)$ for a GNN model, the performance approximation error $\mathcal{O}_\mathcal{W}$ of condensed graph is jointly bounded by the reconstruction term $\mathcal{O}_{\mathcal{Z}}$ and the fitting term $\mathcal{O}_{\mathcal{Y}}$.
\begin{equation}\label{eq_z_gc}\small
 \underbrace{\|\mathbf{W} - \tilde{\mathbf{W}}\|_F}_{\mathcal{O}_{\mathcal{W}}} \leq \|\boldsymbol{\mathcal{M}}_{\mathcal{Z}}(\mathbf{Y})\|_F \cdot \underbrace{\|\boldsymbol{\mathcal{M}}_{\mathcal{Z}}(g(\mathbf{L})\mathbf{X})^\dagger-(g(\tilde{\mathbf{L}})\tilde{\mathbf{X}})^\dagger\|_F}_{\mathcal{O}_{\mathcal{Z}}} + \underbrace{\|\boldsymbol{\mathcal{M}}_{\mathcal{Z}}(\mathbf{Y})-\tilde{\mathbf{Y}}\|_F}_{\mathcal{O}_{\mathcal{Y}}} \cdot  \|(g(\tilde{\mathbf{L}})\tilde{\mathbf{X}})^\dagger\|_F
\end{equation}
where $\boldsymbol{\mathcal{M}}_{\mathcal{Z}}(\cdot)$ is any mapping function that aligns $g(\tilde{\mathbf{L}})\tilde{\mathbf{X}}$ and $g(\mathbf{L})\mathbf{X}$ or $\tilde{\mathbf{Y}}$ and $\mathbf{Y}$. $\mathbf{L}=\mathbf{I}_N-\mathbf{A}$ and $\tilde{\mathbf{L}}=\mathbf{I}_M-\tilde{\mathbf{A}}$ are the Laplacian matrices of $\mathcal{G}$ and $\tilde{\mathcal{G}}$.
\end{proposition}
\vspace{-0.2em}
Proposition \ref{proposition_1} reveals that the reconstruction term $\mathcal{O}_{\mathcal{Z}}$ and the fitting term $\mathcal{O}_{\mathcal{Y}}$ are the two key factors for ideal GC.
Unfortunately, current GC methods rely on specific GNNs (i.e. the rigid filter $g(\cdot)$), resulting in spectrally limited condensed graphs with higher reconstruction errors for $\mathcal{O}_\mathcal{Z}$.
Moreover, the pre-defined task signal in condensation restricts $\mathcal{O}_{\mathcal{Y}}$ to the specific task. Consequently, the condensed graph can only achieve satisfactory performance with the pre-defined GNNs and task, severely limiting its generalizability and reusability. To overcome these limitations, we focus on proposition \ref{proposition_1} and relax the above conditionally constrained GC objective into a generalized one:

\begin{definition}\label{ugc} \textbf{Generalized Graph Condensation.} 
Given an original graph $\mathcal{G}$, the objective of ideal graph condensation is to seek a condensed graph $\tilde{\mathcal{G}}$ that simultaneously  minimizes the representation-level and the semantic-level discrepancies relative to $\mathcal{G}$:
\begin{equation}\label{eq_unify_gc}
\tilde{\mathcal{G}}^* = \arg \underset{\tilde{\mathcal{G}}}{\min} \big\{\underbrace{\Delta_{\mathcal{Z}}({\mathbf{Z}, \tilde{\mathbf{Z}}})}_{\mathcal{O}_{\mathcal{Z}}} + \xi \underbrace{\Delta_\mathcal{Y}(\mathcal{G},\tilde{\mathcal{G}})}_{\mathcal{O}_{\mathcal{Y}}}\big\}
\end{equation}
\vspace{-0.1cm}
where $\Delta_{\mathcal{Z}}(\cdot, \cdot)$ measures the representation-level discrepancy between representations $\mathbf{Z}$ and $\tilde{\mathbf{Z}}$ of $\mathcal{G}$ and $\tilde{\mathcal{G}}$. $\Delta_\mathcal{Y}(\cdot,\cdot)$ is a distance function measuring semantic-level discrepancy, and  $\xi$ serves as a trade-off parameter balancing the fitting and reconstruction terms.

\end{definition} 

Definition \ref{ugc} presents a general GC optimization paradigm, of which current GC methods can be regarded as variant forms (Appendix C.2 shows more details). In this formulation, the generalized node representation $\mathbf{Z}$ replaces $g(\mathbf{L})\mathbf{X}$, and the task-specific fitting term $\mathcal{O}_\mathcal{Y}$ is relaxed to semantic alignment between two graphs.
Therefore, by boosting the diversity of the representations $\mathbf{Z}$ and $\tilde{\mathbf{Z}}$, and designing unsupervised signals for $\Delta \mathcal{Y}$, it is possible to readily establish a pre-trained graph condensation framework. 
Such a framework enhances the generalization of the condensed graph and augments its adaptability across diverse scenarios.

\vspace{-0.1em}
\section{Methodology}
\vspace{-0.1em}

\begin{figure}[htbp!]
\centering
\includegraphics[width=5.0in]{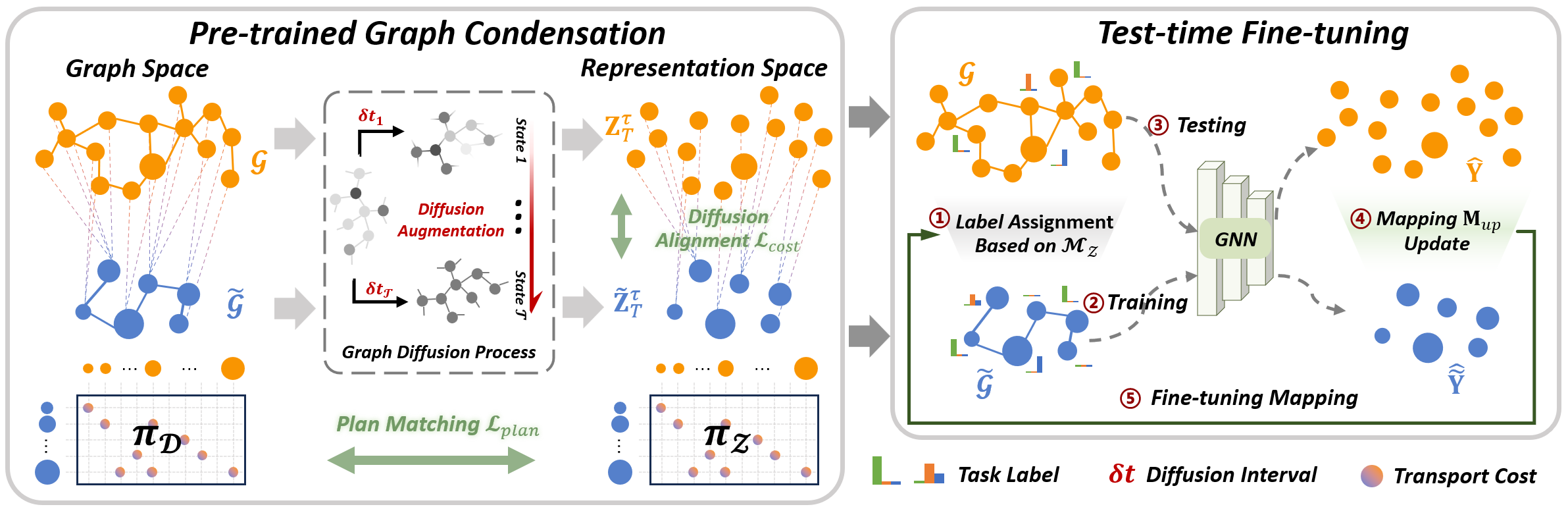}
\caption{Overall pipeline of the proposed PGC framework. $\pi_\mathcal{D}$ and $\pi_\mathcal{Z}$ denote optimal transport plans for graph alignment and for representation alignment, respectively.}
\label{fig_pgc_framework}
\vspace{-0.3cm}
\end{figure}

Fig. \ref{fig_pgc_framework} illustrates the framework of PreGC, which strictly follows the GC objective in Eq. \eqref{eq_unify_gc}. In pre-training phase, graph diffusion augmentation is proposed to increase the diversity of representations in $\mathcal{O}_\mathcal{Z}$ to alleviate architectural dependencies. Subsequently, to realize task-agnostic graph condensation, PreGC adopts an innovative transport plan matching to align semantic associations in $\mathcal{O}_\mathcal{Y}$. In addition, test-time fine-tuning further achieves dynamic adaptation on specific tasks and architectures.

\subsection{Diffusion Augmentation via Graph Heat Equation}\label{Graph_Diffusion_Augmentation}
As analyzed in Section \ref{revisiting}, most GC methods adopt $K$-order SGC \cite{sgc_icml19} or GCN \cite{gcn_iclr17} for encoding graphs to obtain $\mathbf{Z}$ and $\tilde{\mathbf{Z}}$ \cite{gcbench_nips24}.
However, parameters in neural networks may inherit task-induced biases, thus distorting the intrinsic representation properties. 
Although GDEM \cite{gdem_icml24} and CGC \cite{cgc_www25} attempt to use parameter-free encoders to obtain node representations, they remain architecture-bound, resulting in condensed graphs hard to generalize to other GNNs. More critically, under discrete propagation, the representation distributions of two graphs evolve at markedly divergent rates due to diameter disparities, inevitably exacerbating the difficulty of GC.

The graph heat equation (GHE) \cite{chung1997spectral}, as a generalization of the diffusion equation on graph data (graph diffusion), serves as an effective solution to the above problems due to its non-parametric and dynamic continuity properties.
Formally, GHE can be defined as $\frac{d\mathbf{Z}_t}{dt} = -\mathbf{L}\mathbf{Z}_t$, where $\mathbf{Z}_t$ is the evolved input representation at time $t$, and $\mathbf{Z}_0=\mathbf{X}$.
In our proposed PreGC, graph diffusion \cite{diffusion_aaai24} is used to encode node representations and solved using Explicit Euler method:
\begin{equation}\label{eq_numerical_ghe}
\mathbf{Z}_{t+\delta t} = \mathbf{Z}_t - \delta t \mathbf{L} \mathbf{Z}_t = \mathbf{A}^{(\delta t)}\mathbf{Z}_t
\end{equation}
where $\delta t$ is diffusion interval and $\mathbf{A}^{(\delta t)} = (1-\delta t)\mathbf{I} + \delta t \mathbf{A}$. Finally, the representation (or state) $\mathbf{Z}_T$ at the terminal time $T=K\cdot \delta t$ can be formulated as $\mathbf{Z}_T=[\mathbf{A}^{(\delta t)}]^K\mathbf{X}$.
Adjusting $\delta t$ enables smoother propagation rate modulation, mitigating representation instability in discrete propagation.

\textbf{Graph Diffusion Augmentation.} 
Indeed, deterministic diffusion states capture specific graph spectral responses. 
To eliminate architectural bias, the condensed graph should maintain state consistency with the original graph for arbitrary diffusion times to inherit the complete spectral properties.

As a sample generation strategy, data augmentation \cite{DBLP:conf/icde/CuiCYDF024, DBLP:journals/apin/LiLYZZ23} aims to enhance the generalizability and robustness of the model through diverse samples. 
Inspired by this, a simple yet efficient diffusion augmentation is proposed to improve the generalizability of the condensed graph by generating diverse diffusion states with hybrid intervals.
This process can be formulated as:
\begin{equation}\label{eq_diffusion_aug1}
\mathcal{S}(\mathbf{Z}_{T}^\tau) = \big\{\mathbf{Z}^\tau_T ~|~\mathbf{Z}^\tau_T = [\mathbf{A}^{(\delta t_{\tau})}]^{K}\mathbf{X}, ~ \delta t_\tau \sim \mathcal{P}, ~\forall \delta t_\tau \le \frac{2}{\lambda_{\text{max}}(\mathbf{L})} \big\}_{\tau=0}^{\mathcal{T}}
\end{equation}
where $\mathcal{P}=\mathcal{U}(\delta t_{\text{min}},{2}/{\lambda_{\text{max}}(\mathbf{L})})$), $\mathcal{T}$ denotes the condensation epoch, and $\lambda_{\text{max}}(\mathbf{L})$ is the maximum eigenvalue of $\mathbf{L}$. $\forall \delta t_\tau \le {2}/{\lambda_{\text{max}}(\mathbf{L})}$ ensures the stability of the graph diffusion. Similarly, $\mathcal{S}(\tilde{\mathbf{Z}}_T^\tau)$ is obtained by keeping the same $\delta t_\tau$ as above on $\tilde{\mathcal{G}}$. 

It can be seen from Eq~\eqref{eq_diffusion_aug1} that the stochastic variation of the diffusion interval $\delta t$ raises uncertainty of the diffusion state at time $T$. By aligning the two graph representations ${\mathbf{Z}}_T^\tau$ and $\tilde{\mathbf{Z}}_T^\tau$, the condensed graph is guided to capture the diffusion trajectories of the original graph across varying states. 
In addition, we can derive the following proposition (Proofs are in Appendix A): 
\begin{proposition}\label{diffusion_proposition}
When enough diffusion times $\{\delta t_\tau\}_{\tau=1}^{\mathcal{T}}$ are sampled from the distribution $\mathcal{P}$ over the interval $I=[\delta t_{\min}, 2/\lambda_{\max}(\mathbf{L})]$, the spectral response function is satisfied by sampling coverage over the entire spectral range:
\begin{equation}
\lim_{\mathcal{T}\rightarrow \infty}\mathrm{Pr}(\max_{i}\sup_{\delta t \in I} \min_{\tau}|\Phi_i(\delta t)-\Phi_i(\delta t_{\tau})|<\varphi)=1, ~\forall \varphi>0
\end{equation}
where $\Phi_i(\delta t)=e^{-K\delta t{\lambda_i}}$ is the spectral response function for the $i$-th eigenvalue $\lambda_i$.
\end{proposition}
\vspace{-0.1cm}
Proposition \ref{diffusion_proposition} establishes a fundamental probabilistic guarantee for the completeness of spectral coverage in graph diffusion augmentation. This strategy breaks through the fixed-architecture constraints in conventional GC methods and effectively preserves the original spectral properties.

\subsection{Optimal Transport Plan Matching} \label{Optimal_Transport_Plan_Matching}
For the generalized GC objective in Eq .\eqref{eq_unify_gc}, the distance function $\Delta_{\mathcal{Y}}(\cdot,\cdot)$ is challenging to define without any task signal.
However, semantic information should be an intrinsic property of the graph, rather than being influenced by the downstream task. In other words, the semantics of the same node should remain unchanged in both the graph and the representation spaces. 
This insight inspired us to ensure the consistency of semantic associations between $\mathcal{G}$ and $\tilde{\mathcal{G}}$ by defining two similarity metrics in graph and representation spaces, respectively. In fact, it is nontrivial to directly compare these two metrics, since they belong to two different spaces and are two disjoint objects by nature.

Fortunately, optimal transport theory provides an elegant framework for this purpose. Therefore, we present an optimal transport plan matching, which aims to find the consistent optimal plan for graph alignment and representation alignment.
For the graph space, a natural idea is to consider both node features and structure information in the transport plan. In PreGC, the fused Gromov-Wasserstein \cite{FGWD_algorithms20} is adopted to obtain the optimal transport plan $\pi^{*}_\mathcal{D} \in \mathbb{R}^{N\times M}$:
\begin{equation}\label{eq_g_plan}
\pi^{*}_\mathcal{D}(\mathcal{G},\tilde{\mathcal{G}})=\mathop{\rm arg~min} \limits_{\pi_{\mathcal{D}}\in{\Pi}(\mu,\nu)} \big\langle \gamma\boldsymbol{\mathcal{K}}(\mathbf{X},\tilde{\mathbf{X}})+(1-\gamma)\boldsymbol{\mathcal{J}}(\mathbf{A},\tilde{\mathbf{A}})\otimes\pi_\mathcal{D},\pi_\mathcal{D}  \big\rangle
\vspace{-0.2em}
\end{equation}
where $\otimes$ denotes the tensor-matrix multiplication, and $\gamma$ represents the coefficient used to control the importance of structure and features. $\boldsymbol{\mathcal{K}}(\mathbf{X},\tilde{\mathbf{X}})_{i,j}=\|\mathbf{X}_i-\tilde{\mathbf{X}}_j\|_2$ and $\boldsymbol{\mathcal{J}}(\mathbf{A},\tilde{\mathbf{A}})_{ij,kl}=|\mathbf{A}_{ik}-\tilde{\mathbf{A}}_{jl}|$ are cost metric functions that measure feature discrepancy and structure discrepancy, respectively.
For the node presentation, we leverage Wasserstein distance to obtain the optimal plan $\pi^*_\mathcal{Z} \in \mathbb{R}^{N \times M}$:
\begin{equation}\label{eq_z_plan}
\pi^{*}_\mathcal{Z}(\mathbf{Z}^\tau_T,\tilde{\mathbf{Z}}^\tau_T)=\mathop{\rm arg~min} \limits_{\pi_{\mathcal{Z}}\in{\Pi}(\mu,\nu)} \big\langle {\boldsymbol{\mathcal{K}}}(\mathbf{Z}^\tau_T,\tilde{\mathbf{Z}}^\tau_T),\pi_\mathcal{Z}  \big\rangle
\end{equation}
Intuitively, the optimal transport plan characterizes the transferring probability from source nodes to condensed nodes, and semantic association consistency in different spaces is preserved by minimizing the divergence between two transport plans. The transport plan matching loss is defined as follows:
\begin{equation}\label{eq_loss_plan}
\mathcal{L}_{plan}=\Delta_{\mathcal{Y}}\big(\pi^{*}_\mathcal{D}(\mathcal{G},\tilde{\mathcal{G}}),\pi^{*}_\mathcal{Z}(\mathbf{Z}_T^\tau,\tilde{\mathbf{Z}}_T^\tau)\big)
\end{equation}
Without loss of generality, $\Delta_{\mathcal{Y}}(\cdot,\cdot)$ is set to the Frobenius norm in this paper.

\subsection{Node Significance Evaluation}\label{node_sig_eva}
Unlike conventional GC works without explicitly considering node contributions for condensation, the optimal resresentation transport plan $\pi^*_\mathcal{Z}$ here not only delineates the semantic associations between both graphs, but also inversely reflects the potential significance of the source nodes. Specifically, let's first define a mask matrix $\pi_{mask}\in \mathbb{R}^{N\times M}$ to
filter out the most influencial nodes from source graph $\mathcal{G}$ on each condensed node $j$ , i.e., we have  $(\pi_{mask})_{i,j} = (\pi^*_\mathcal{Z})_{i,j}$ when  $(\pi^*_\mathcal{Z})_{i,j}$ is among the top-$H$ of the $j$-th  column, otherwise $(\pi_{mask})_{i,j} = 0$. Then, the node significance score vector $\mathbf{s} \in \mathbb{R}^N$ of the source graph can be obtained by:
\begin{equation}\label{eq:top_s}
\mathbf{s}= \pi_{mask}\cdot \mathbf{1}_M
\end{equation}
where $\mathbf{1}_M \in \mathbb{R}^{M}$ denotes the all-ones vector.
Essentially, the significance  $\mathbf{s}$ explicitly reveals the contributions of different source nodes for the condensed graph. Noticeably, the evaluation of node significance based on $\mathbf{s}$ are two-folds. On the one hand, it provides more transparent interpretability for graph condensation, and  on the other hand, it also helps to guide node labeling effectively in an active learning way \cite{DBLP:conf/iclr/HanLMT0TY24}. Section \ref{Experimental_Results}  further confirms the validity of the node significance evaluation.

\subsection{Condensed Graph Optimization}
\textbf{Pre-training.}
Benefiting from Eq. \eqref{eq_z_plan}, the reconstruction term $\mathcal{O}_\mathcal{Z}$ can be achieved by diffusion alignment between $\mathbf{Z}_T^\tau$ and $\tilde{\mathbf{Z}}_T^\tau$. Therefore, according to Eq. (\ref{eq_unify_gc}), the total loss can be denoted as:  
\begin{equation}\label{eq_loss_cost}
\mathcal{L}_{total}=\mathcal{L}_{cost}+\xi\mathcal{L}_{plan},~~~~\mathcal{L}_{cost}= \Delta_{\mathcal{Z}}(\mathbf{Z}_T^\tau,\tilde{\mathbf{Z}}_T^\tau)
\end{equation}
where $\Delta_{\mathcal{Z}}(\cdot,\cdot)$ denotes the Wasserstein distance \cite{WD_access25}.
After obtaining the condensed graph $\tilde{\mathcal{G}}$, the transport plan $\pi_{\mathcal{Z}}^*$, which reflects the transfer probabilities between the source nodes from $\mathcal{G}$ and the condensed nodes, is also retained. To ensure the uniqueness of the semantic mapping, we discretize $\pi_{\mathcal{Z}}^*$ to obtain the semantic assignment matrix $\mathbf{M} \in \mathbb{R}^{N\times M}$:
\begin{equation}\label{eq_M}
\mathbf{M}_{i,j}=\mathbb{I}(j=  \underset{1\le l \le M}{\arg\max}~(\pi_{\mathcal{Z}}^*)_{i,l})
\end{equation}
%
%
%
%
When the condensed graph $\tilde{\mathcal{G}}$ needs to be trained on downstream tasks, the target variable $\tilde{\mathbf{Y}} \in \mathbb{R}^{M\times C}$ can be obtained from the original label by the traceable semantic harmonizer $\boldsymbol{\mathcal{M}}_{\mathcal{Z}}(\cdot)$:
\begin{equation}\label{eq_condensed_y}
\tilde{\mathbf{Y}} = \boldsymbol{\mathcal{M}}_{\mathcal{Z}}(\mathbf{Y}_{tr}) = \mathbf{D}^{-1}_\Omega\mathbf{M}_{\Omega}^\top\mathbf{Y}_{tr}
\end{equation}
where $\mathbf{Y}_{tr} \in \mathbb{R}^{N_{tr}\times C}$ is a training label with $N_{tr}$ labelled nodes, and $\Omega = \{i_1,...,i_{N_{tr}}\} $ is the set of their row indices. $\mathbf{M}_\Omega \in \mathbb{R}^{N_{tr}\times M}$ denotes the sub-assignment matrix defined by $(\mathbf{M}_\Omega)_{k,j}=\mathbf{M}_{i_k,j}$.  $\mathbf{D}_\Omega=\text{diag}(\mathbf{M}_\Omega^\top\mathbf{1}_{N_{tr}}) \in \mathbb{R}^{M\times M}$, and $\text{diag}(\cdot)$ is used to create a diagonal matrix. 

\textbf{Test Time Fine-tuning.}
To further unlock the potential of condensed graphs while improving the performance of condensed graphs on different tasks and GNN architectures, we draw on the idea of fine-tuning pre-trained models \cite{DBLP:conf/emnlp/ByunCKP24} to propose PreGC$_\text{ft}$, a strategy for achieving finer semantic alignment by fine-tuning the assignment matrix. Specifically, given a decay rate $\epsilon \in [0,1]$, the assignment matrix $\mathbf{M}$ is updated every $\tau_{up}$ epochs according to predefined tasks and GNN:
\begin{equation}\label{eq_condensed_y_sft}
\mathbf{M} \leftarrow \epsilon \mathbf{M} + (1-\epsilon)\mathbf{M}_{up}
\end{equation}
where $\mathbf{M}_{up} \in \mathbb{R}^{N\times M}$ is obtained from $\tilde{\pi}^*_{\mathcal{Y}}(\hat{\mathbf{Y}},\hat{\tilde{\mathbf{Y}}})$ similar as Eq. \eqref{eq_z_plan}. $\hat{\mathbf{Y}}$ and $\hat{\tilde{\mathbf{Y}}}$ are the predicted outputs by the specific GNN encoded from $\mathcal{G}$ and $\tilde{\mathcal{G}}$, respectively.

Through the above optimization framework, we empower GC with pre-training and fine-tuning capabilities. The overall pipeline of PreGC are summarized in Appendix B.

\section{Experiments}
\vspace{-0.1cm}
\subsection{Experimental Settings}\label{Experimental_Settings}
\vspace{-0.3em}
\textbf{Datasets.} To comprehensively evaluate the condensation performance of PreGC, five graph datasets (Cora, Citeseer, Pubmed~\cite{gcn_iclr17}, OGB-Arxiv~\cite{DBLP:conf/kdd/Zhang0Z0Z0025}, and H\&M~\cite{DBLP:journals/corr/abs-2409-14500}) are utilized in experiments. Different tasks are tested to validate the generalization of PreGC, including node classification (NC), node clustering (NClu), link prediction (LP), and node regression (NR). 
Specifically, OGB-Arxiv includes two NC tasks: predicting paper topics (T) and publication years (Y). H\&M includes an NC task: predicting product categories (C), and an NR task: predicting product prices (P).
Two condensation ratios ($r=\frac{M}{N}$) are considered. More details are provided in Appendix C.1.

\textbf{Baselines \& Implementations.}
We compare the proposed PreGC with six representative graph condensation methods (GCDM \cite{gcdm_arxiv22}, GCond \cite{gcond_iclr22}, SFGC \cite{sfgc_nips23}, SGDD \cite{sgdd_nips23}, GDEM \cite{gdem_icml24}, and CGC \cite{cgc_www25}). 
Following \cite{gcond_iclr22, gcbench_nips24}, the condensed graph is synthesized according to the default settings of the above methods, and a certain GNN (default is SGC \cite{sgc_icml19}) is trained on this graph and evaluated on the test set of the original graph.
Moreover, other eight representative GNNs (GCN \cite{gcn_iclr17}, APPNP \cite{appnp_iclr19}, $k$-GNN \cite{kgnn_aaai19}, GAT \cite{gat_iclr18}, GraphSAGE \cite{graphsage_nips17}, SSGC \cite{ssgc_iclr21}, BernNet \cite{bernnet_nips21}, and GPRGNN \cite{gprgnn_iclr21}) are selected to evaluate the effectiveness of PreGC. The details of baselines, task settings, condensation settings, and parameter settings are provided in Appendix C.2 and C.3.

\vspace{-0.2em}
\begin{table}[htbp!]
\tiny
  \centering
  \caption{Performance comparison on different tasks (We report test accuracy (\%) on NC, NMI (\%) on NClu, and AUC (\%) on LP tasks. \textbf{Bold entry} is the best result, \underline{underline} marks the runner-up).}
    \begin{tabular}{ccccccccccc}
    \toprule
    \multirow{2}[2]{*}{\scriptsize{Dataset}} & \multirow{2}[2]{*}{\scriptsize{Task}} & \multirow{2}[2]{*}{\scriptsize{Ratio}} & GCDM  & GCond & SFGC  & SGDD  & GDEM  & CGC   & PreGC   & \multirow{2}[2]{*}{\scriptsize{Whole}} \\
          &       &       & (2022) & (2022) & (2023) & (2023) & (2024) & (2025) & (Our) &  \\
    \midrule
    \multirow{8}[7]{*}{\textbf{\tiny{Cora}}} & \multicolumn{1}{c}{\multirow{2}[2]{*}{\tiny{NC}}} & 1.3\%  & 61.5±1.2 & 79.0±1.3 & 75.4±0.5 & 79.2±0.6 & \underline{80.8±0.4} & 79.0±0.8 & \textbf{81.1±0.3} & \multirow{2}[2]{*}{80.8±0.8} \\
          &       & 2.6\%   & 69.2±0.4 & 79.0±0.6 & 79.2±1.4 & 81.1±0.8 & 80.7±0.1 & \textbf{81.8±0.6} & \underline{81.6±0.9} &  \\
\cmidrule{2-11}          & \multicolumn{1}{c}{\multirow{2}[2]{*}{\tiny{NC→NClu}}} & 1.3\%   & 36.9±0.7 & 58.1±1.6 & 56.0±2.0 & \textbf{59.0±1.2} & 57.2±0.8 & \underline{58.2±1.2} & 54.6±1.0 & \multirow{2}[2]{*}{59.3±1.5} \\
          &       & 2.6\%   & 45.3±0.5 & 57.6±0.9 & 57.3±0.7 & 58.7±0.5 & 57.0±0.5 & \underline{59.4±0.5} & \textbf{59.6±0.5} &  \\
\cmidrule{2-11}          & \multicolumn{1}{c}{\multirow{2}[2]{*}{\tiny{LP}}} & 1.3\%   & 56.8±0.7 & 56.6±3.0 & N/A     & 57.0±3.5 & \underline{66.9±0.6} & 65.5±0.8 & \textbf{75.3±2.2} & \multirow{2}[2]{*}{81.9±4.8} \\
          &       & 2.6\%   & 57.2±1.8 & \underline{62.2±1.6} & N/A     & 60.4±2.2 & 59.7±2.6 & 46.5±4.1 & \textbf{79.2±2.2} &  \\
\cmidrule{2-11}          & \multicolumn{1}{c}{\multirow{2}[1]{*}{\tiny{LP→NClu}}} & 1.3\%   & 19.1±4.7 & 26.8±10.1 &  N/A     & 14.3±6.6 & \underline{29.5±3.8} & 8.9±0.7 & \textbf{33.6±1.0} & \multirow{2}[1]{*}{42.3±3.3} \\
          &       & 2.6\%   & 13.1±2.8 & 32.0±6.0 & N/A     & 13.4±3.3 & \textbf{35.2±1.4} & 10.3±1.9 & \underline{34.9±2.1} &  \\
\cmidrule{1-11}     \multirow{8}[6]{*}{\textbf{\tiny{Citeseer}}} & \multicolumn{1}{c}{\multirow{2}[1]{*}{\tiny{NC}}} & 0.9\%  & 54.7±1.2 & 68.9±1.0 & 67.1±0.9 & 68.9±1.6 & \textbf{70.6±0.3} & \underline{70.5±0.4} & \textbf{70.6±0.5} & \multirow{2}[1]{*}{69.2±0.4} \\
          &       & 1.8\%   & 52.1±0.9 & 67.9±0.7 & 65.1±1.5 & 69.7±1.7 & 70.5±0.4 & \underline{70.6±0.4} & \textbf{70.8±0.8} &  \\
\cmidrule{2-11}          & \multicolumn{1}{c}{\multirow{2}[2]{*}{\tiny{NC→NClu}}} & 0.9\%   & 25.3±0.7 & 41.0±2.7 & 42.4±1.8 & \underline{43.0±1.3} & 42.8±0.7 & 40.6±1.7 & \textbf{43.8±1.4} & \multirow{2}[2]{*}{42.4±0.5} \\
          &       & 1.8\%   & 20.8±2.5 & 38.1±0.8 & 42.9±0.9 & 43.5±0.7 & \textbf{45.2±0.4} & 40.7±1.8 & \underline{43.9±0.6} &  \\
\cmidrule{2-11}          & \multicolumn{1}{c}{\multirow{2}[2]{*}{\tiny{LP}}} & 0.9\%   & 61.5±3.8 & 61.5±2.0 & N/A     & 60.0±2.2 & \underline{65.8±3.7} & N/A     & \textbf{66.5±1.6} & \multirow{2}[2]{*}{80.8±4.9} \\
          &       & 1.8\%   & 57.5±5.3 & 62.3±3.4 & N/A     & 56.9±2.7 & \underline{64.3±3.6} & 55.5±3.9 & \textbf{69.3±3.3} &  \\
\cmidrule{2-11}          & \multicolumn{1}{c}{\multirow{2}[1]{*}{\tiny{LP→NClu}}} & 0.9\%   & 7.2±0.9 & 20.3±3.3 & N/A     & 10.1±1.2 & \underline{21.0±2.5} & N/A     & \textbf{32.7±0.7} & \multirow{2}[1]{*}{33.4±3.8} \\
          &       & 1.8\%   & 9.3±3.0 & 14.7±3.0 & N/A     & 9.2±1.4 & \underline{16.9±3.3} & 7.2±1.4 & \textbf{33.1±1.2} &  \\
\cmidrule{1-11}     \multirow{8}[7]{*}{\textbf{\tiny{Pubmed}}} & \multicolumn{1}{c}{\multirow{2}[1]{*}{\tiny{NC}}} & 0.09\%   & 71.4±1.6 & 72.9±0.6 & 77.4±0.2 & 78.4±0.4 & 77.8±0.1 & \underline{78.7±0.3} & \textbf{79.9±0.4} & \multirow{2}[1]{*}{78.8±0.2} \\
          &       & 0.15\%   & 60.8±0.8 & 71.6±1.4 & 78.1±0.4 & \underline{78.6±0.5} & 76.8±0.6 & 78.4±0.5 & \textbf{81.0±0.5} &  \\
\cmidrule{2-11}          & \multicolumn{1}{c}{\multirow{2}[2]{*}{\tiny{NC→NClu}}} & 0.09\%   & 24.8±2.8 & 21.1±5.5 & 35.3±0.6 & \underline{35.6±0.8} & 35.5±0.1 & \textbf{37.0±0.6} & \underline{35.6±0.8} & \multirow{2}[2]{*}{37.1±0.8} \\
          &       & 0.15\%   & 19.4±0.5 & 18.9±0.7 & 36.9±0.2 & 36.1±0.3 & 34.1±0.4 & \underline{38.0±0.4} & \textbf{38.3±0.7} &  \\
\cmidrule{2-11}          & \multicolumn{1}{c}{\multirow{2}[2]{*}{\tiny{LP}}} & 0.09\%   & 72.1±2.9 & \underline{72.9±3.8} & N/A     & \underline{72.9±3.1} & 43.5±1.6 & N/A     & \textbf{82.9±2.0} & \multirow{2}[2]{*}{95.0±1.0} \\
          &       & 0.15\%   & 68.9±4.2 & 73.7±3.5 & N/A     & 68.4±3.2 & 56.1±2.3 & \underline{84.8±2.0} & \textbf{86.3±2.4} &  \\
\cmidrule{2-11}          & \multicolumn{1}{c}{\multirow{2}[2]{*}{\tiny{LP→NClu}}} & 0.09\%  & \underline{14.1±3.0} & 6.2±4.4 & N/A     & 3.0±0.4 & 1.6±0.2 & N/A     & \textbf{22.5±0.7} & \multirow{2}[2]{*}{30.7±3.3} \\
          &       & 0.15\%   & 13.7±0.4 & 4.1±1.8 & N/A     & 12.0±1.8 & 20.1±4.2 & \underline{23.5±0.3} & \textbf{24.0±1.0} &  \\
    \bottomrule
    \end{tabular}%
  \label{tab:main_results_1}%
\end{table}%

\vspace{-0.25cm}
\subsection{Experimental Results}\label{Experimental_Results}
\vspace{-0.05cm}
\textbf{Performance of PreGC on Various Tasks.} 
We first demonstrate the generalizability of the condensed graph obtained by PreGC as shown in Table \ref{tab:main_results_1}. Clearly, PreGC achieves optimal performance in most cases.
In particular, PreGC consistently outperforms baselines on LP, demonstrating that the condensed graph of PreGC can well capture the topological relationships in the original structure.
%
%
%
%
%
SFGC is unable to perform LP task (i.e. "N/A") due to its emphasis on structure-free during condensation, limiting its task reusability.
In addition, Table \ref{tab:main_results_2} shows the generalization of PreGC under different supervision tasks and the flexibility in task migration.
Owing to the limitations that need to initialize condensed labels by the original class labels, existing methods can't be directly applied to node regression (i.e. "N/A"). By virtue of its unsupervised matching strategy, PreGC becomes the first GC method capable of adapting to NR task. Regardless of the tasks, PreGC consistently achieved optimal performance. 
In addition, the fine-tuning strategy PreGC$_{\text{ft}}$ achieves 1.46\% gain compared to PreGC on H\&M dataset. 
\vspace{-0.2cm}
%
%

%
\begin{table}[htbp!]\tiny
  \centering
  \caption{Performance comparison to baselines on different supervised tasks ("Y→T" means condensation on the "Y" task and testing on the "T" task. We report test accuracy (\%) on NC and R$^2$ (\%) on NR tasks. \textcolor[rgb]{ .753,  0,  0}{\textbf{Imp.}} indicates the performance improvement over the best baseline).}
    \begin{tabular}{cccccccccc}
    \toprule
          & \multicolumn{4}{c}{OGB-Arxiv ($r=1.25\%$)} &       & \multicolumn{4}{c}{H\&M ($r=2.5\%$)} \\
\cmidrule{2-5}\cmidrule{7-10}          & T→T   & Y→T   & Y→Y   & T→Y   &       & C→C   & P→C   & P→P   & C→P \\
\cmidrule{2-5}\cmidrule{7-10}    \rowcolor[rgb]{ .949,  .949,  .949} Whole & \multicolumn{2}{c}{64.43 ± 0.24} & \multicolumn{2}{c}{55.37 ± 0.21} &       & \multicolumn{2}{c}{77.07 ± 0.26} & \multicolumn{2}{c}{50.80 ± 0.64} \\
    GCDM  & 33.08±1.04 & 46.48±2.40 & 44.20±0.84 & 46.17±1.25 &       & 57.15±0.90 & N/A     & N/A     & 26.08±0.76 \\
    GCond & 56.74±0.65 & 41.38±1.65 & 49.25±0.67 & 42.52±2.12 &       & 64.57±0.53 & N/A     & N/A     & 15.29±1.14 \\
    SFGC  & 59.64±0.22 & 50.15±1.55 & 50.07±0.47 & 44.30±1.76 &       & 60.09±1.24 & N/A     & N/A     & 4.81±1.04 \\
    SGDD  & 58.68±0.82 & 27.80±2.09 & 41.70±2.25 & 42.30±1.58 &       & 61.85±0.39 & N/A     & N/A     & 18.85±2.13 \\
    GDEM  & 53.93±0.71 & 40.01±0.87 & 49.17±0.32 & 47.80±0.96 &       & 52.61±3.05 & N/A     & N/A     & 6.83±2.99 \\
    CGC   & 58.71±0.65 & 53.32±1.61 & 49.50±0.72 & 48.53±0.24 &       & 64.35±0.33 & N/A     & N/A     & 15.95±1.11 \\
    \midrule
    \textbf{PreGC}   & \underline{60.55±0.43} & \underline{59.62±0.47} & \underline{52.34±0.30} & \underline{51.49±0.36} &       & \underline{69.06±0.25} & \underline{68.86±0.29} & \underline{34.39±0.93} & \underline{33.17±1.05} \\
    \textcolor[rgb]{ .753,  0,  0}{\textbf{Imp.}} & \textcolor[rgb]{ .753,  0,  0}{0.91↑} & \textcolor[rgb]{ .753,  0,  0}{6.30↑} & \textcolor[rgb]{ .753,  0,  0}{2.84↑} & \textcolor[rgb]{ .753,  0,  0}{2.96↑} & \textcolor[rgb]{ .753,  0,  0}{} & \textcolor[rgb]{ .753,  0,  0}{4.49↑} & -     & -     & \textcolor[rgb]{ .753,  0,  0}{7.09↑} \\
    \midrule
    \textbf{PreGC}$_{\text{ft}}$ & \textbf{60.81±0.40} & \textbf{60.86±0.43} & \textbf{52.47±0.20} & \textbf{52.41±0.25} &       & \textbf{69.98±0.24} & \textbf{69.98±0.28} & \textbf{34.75±0.74} & \textbf{34.63±0.69} \\
    \textcolor[rgb]{ .753,  0,  0}{\textbf{Imp.}} & \textcolor[rgb]{ .753,  0,  0}{1.17↑} & \textcolor[rgb]{ .753,  0,  0}{7.54↑} & \textcolor[rgb]{ .753,  0,  0}{2.97↑} & \textcolor[rgb]{ .753,  0,  0}{3.88↑} & \textcolor[rgb]{ .753,  0,  0}{} & \textcolor[rgb]{ .753,  0,  0}{5.41↑} & -     & -     & \textcolor[rgb]{ .753,  0,  0}{8.55↑} \\
    \bottomrule
    \end{tabular}%
  \label{tab:main_results_2}%

\end{table}%

\vspace{-0.3cm}
\textbf{Generalizability of PreGC across Different GNNs.} Ideal condensed graphs should perform well in different GNNs, and to this end, each synthetic graph is evaluated by nine GNNs, and the results are shown in Table \ref{tab:different_gnns}. It is evident that the PreGC exhibits the highest average accuracy, suggesting that the condensed graph extracted by PreGC can consistently benefit various GNNs. Furthermore, only SFGC and PreGC have average accuracy exceeding that of training on the raw graph. However, SFGC can lead to limited application scenarios due to its lack of explicit structure. 

\vspace{-0.1cm}
%
\begin{table}[htbp!]\tiny
  \centering
  \caption{Node classification performance across different GNNs on Pubmed ($r=0.15\%$). \colorbox[rgb]{ .851,  .851,  .851}{Highlight} marks the lossless results and \textcolor[rgb]{ .753,  0,  0}{\textbf{Avg.}} is the average results of GNNs.}
    \begin{tabular}{ccccccccccc}
    \toprule
     & SGC   & GCN   & APPNP & $k$-GNN & GAT   & SAGE & SSGC  & Bern. & GPR. & \textcolor[rgb]{ .753,  0,  0}{\textbf{Avg.}} \\
    \midrule
    \rowcolor[rgb]{ .949,  .949,  .949} \tiny{Whole}   & 78.9±0.2 & 78.2±0.4 & 79.2±0.4 & 77.0±0.8 & 77.2±0.4 & 77.0±0.5 & 79.0±0.3 & 77.4±0.5 & 79.0±0.3 & 78.1±1.0 \\
    \tiny{GCDM}  & 60.8±0.8 & 56.8±1.1 & 70.0±0.9 & 72.3±0.5 & 76.2±0.8 & 53.3±0.8 & 69.8±0.4 & 68.2±1.3 & 71.1±0.7 & 66.5±7.7 \\
    \tiny{GCond} & 71.6±1.6 & 73.1±1.7 & 73.0±0.2 & 65.9±2.4 & 71.7±2.9 & 69.4±1.5 & 73.3±0.4 & 72.6±2.2 & 72.4±0.8 & 71.4±2.4 \\
    \tiny{SFGC}  & 78.1±0.4 & \cellcolor[rgb]{ .851,  .851,  .851}78.9±0.7 & \cellcolor[rgb]{ .851,  .851,  .851}79.7±0.6 & \cellcolor[rgb]{ .851,  .851,  .851}78.1±0.7 & \cellcolor[rgb]{ .851,  .851,  .851}77.5±0.8 & 75.9±0.5 & 78.7±0.3 & \cellcolor[rgb]{ .851,  .851,  .851}78.4±0.4 & 79.0±0.2 & \cellcolor[rgb]{ .851,  .851,  .851}78.3±1.1 \\
    \tiny{SGDD}  & 78.6±0.5 & \cellcolor[rgb]{ .851,  .851,  .851}78.7±0.3 & 78.7±0.7 & 76.7±0.7 & 73.4±3.0 & 76.4±0.4 & 78.6±0.2 & 77.3±0.5 & 77.4±0.7 & 77.3±1.7 \\
    \tiny{GDEM}  & 76.8±0.6 & 77.3±0.3 & 77.7±0.4 & \cellcolor[rgb]{ .851,  .851,  .851}78.2±0.5 & 76.7±0.7 & 76.4±1.3 & 77.5±1.1 & 76.9±0.4 & 77.0±0.3 & 77.2±0.6 \\
    \tiny{CGC}   & 78.4±0.5 & \cellcolor[rgb]{ .851,  .851,  .851}79.5±0.3 & 78.8±0.4 & 75.7±0.9 & \cellcolor[rgb]{ .851,  .851,  .851}\textbf{78.9±0.5} & 76.4±0.7 & 79.0±0.4 & 77.3±0.4 & 78.5±0.6 & 78.1±1.3 \\
    \tiny{\textbf{PreGC}}   & \cellcolor[rgb]{ .851,  .851,  .851}\textbf{81.0±0.5} & \cellcolor[rgb]{ .851,  .851,  .851}\textbf{80.3±0.4} & \cellcolor[rgb]{ .851,  .851,  .851}\textbf{80.9±0.3} & \cellcolor[rgb]{ .851,  .851,  .851}\textbf{79.1±0.5} & \cellcolor[rgb]{ .851,  .851,  .851}77.8±1.0 & \cellcolor[rgb]{ .851,  .851,  .851}\textbf{77.9±1.0} & \cellcolor[rgb]{ .851,  .851,  .851}\textbf{81.3±0.4} & \cellcolor[rgb]{ .851,  .851,  .851}\textbf{80.2±0.3} & \cellcolor[rgb]{ .851,  .851,  .851}\textbf{80.0±0.5} & \cellcolor[rgb]{ .851,  .851,  .851}\textbf{79.8±1.3} \\
    \bottomrule
    \end{tabular}%
  \label{tab:different_gnns}%
\end{table}%
%
%

%
%
\textbf{Performance Gap between Condensed Graph and Original Graph.} According to \cite{cgc_www25, simgc_pkdd24, gdem_icml24}, the quality of condensed graphs can be quantified through class-level features on NC task.
Therefore, we present the labeled reconstruction error $LRE = \frac{1}{K} \sum_{k=0}^K\| \mathbf{Y}^{\dagger}\mathbf{A}^k\mathbf{X} - \tilde{\mathbf{Y}}^{\dagger}\tilde{\mathbf{A}}^k\tilde{\mathbf{X}} \|_F $ with the class label to evaluate condensed graphs. This formulation systematically 
captures multiscale receptive fields
\begin{wrapfigure}{r}{7.7cm}
    \centering
    \includegraphics[width=0.55\textwidth]{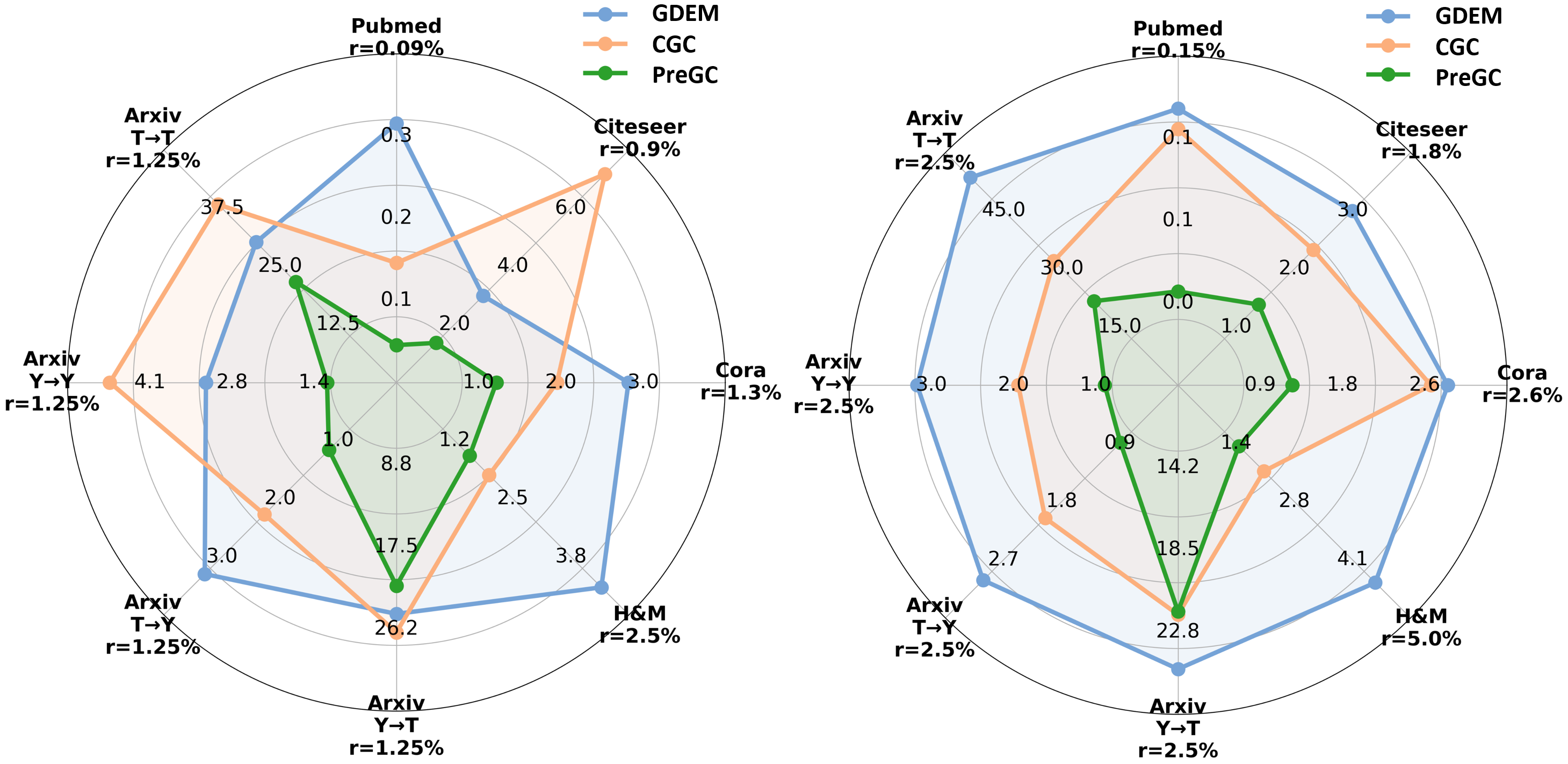}
   
\caption{The $LRE$ of different GC methods.}
    \label{fig_lre}
    
\end{wrapfigure}
through the expectation over propagation 
steps to evaluate the distribution discrepancy.
We set $K = 5$ and the case $k = 0$ is also taken into account, which reflects the difference between the condensed and original features.
As shown in Fig. \ref{fig_lre}, compared with two representative GC methods \cite{cgc_www25, gdem_icml24}, the $LRE$ of PreGC is consistently lower than that of the other two methods on different data and tasks. Since PreGC does not use any task labels in condensation, this is more challenging than baselines and also demonstrates the effectiveness of traceable semantic harmonizer.

\begin{figure}[htbp]
\centering
\includegraphics[width=4.6in]{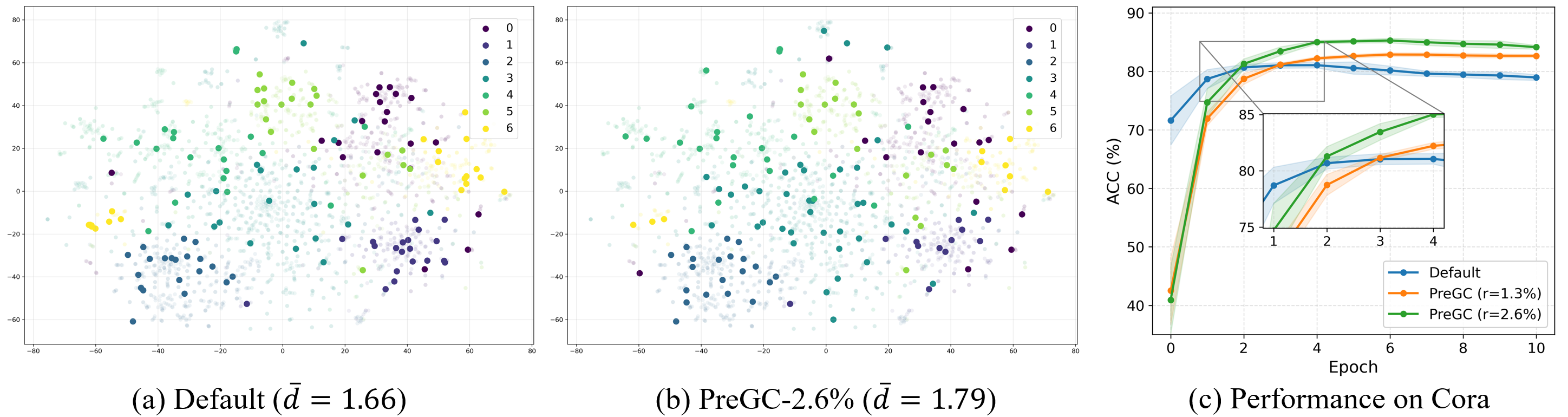}
\caption{Performance of the training set selected via node significance on Cora.}
\label{fig_data_value}
\end{figure}

\textbf{Data Valuation based on Node Significance.} 
To confirm the validity of the significance evaluation in Section \ref{node_sig_eva}, we rescreened the same number of nodes as the original training set based on Eq. \eqref{eq:top_s} and retrained them on the original graph. Fig. \ref{fig_data_value} (b) shows a visualization of the distribution of nodes in the rescreened training set (Dark nodes denote the training set, and distinct colors denote different classes). Compared to the default training set provided by \cite{gcn_iclr17}  (Fig. \ref{fig_data_value} (a)), the rescreened training set demonstrates greater coverage and distributional dispersion. We further measure distribution differences with the average nearest neighbor distance $\bar{d}=\frac{1}{|\mathcal{V}_{tr}|}\sum_{i \in \mathcal{V}_{tr}} \underset{j \neq i}{\min} \| \mathbf{z}_i - \mathbf{z}_j \|_2$ (larger $\bar{d}$ indicates wider coverage of the selected labels, and 2-order diffusion is considered, i.e., $\mathbf{Z} = \mathbf{A}^2\mathbf{X}$ by default), and the results also confirmed the above viewpoint.
Fig. \ref{fig_data_value} (c) compares the performance of SGC trained with different training sets. Leveraging the rescreened nodes as supervision yields higher classification accuracy and markedly improves the training efficiency of the source graph.

\begin{wrapfigure}{r}{5.75cm}

    \centering
    \includegraphics[width=0.35\textwidth]{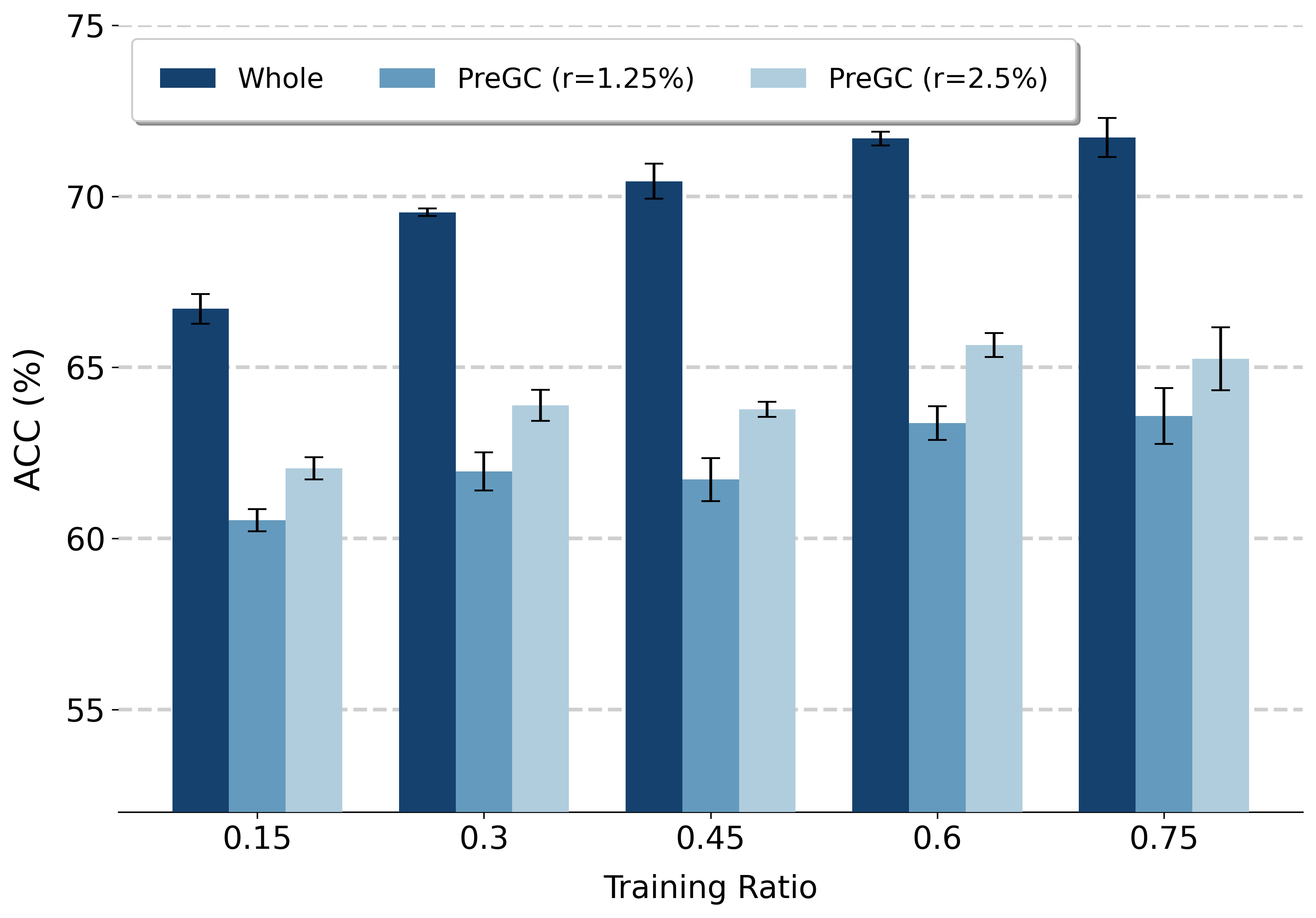}

\caption{The prerformance of PreGC with different trainig ratio on OGB-Arxiv.}
    \label{fig_flexibility}

\end{wrapfigure}
\textbf{Flexibility of PreGC.} 
The existing condensation methods are limited to concentrating under certain proportions of labels. When new labels are added in the original graph, these methods must re-condensation to capture this new knowledge in the condensed graph, which undoubtedly increases the condensation cost and limits its flexibility. PreGC can quickly transfer new labels to the condensed graph through the traceable semantic harmonizer without re-condensation. As shown in Fig. \ref{fig_flexibility}, 
we rescaled the training set ratio of the original graph and assigned labels to the condensed graph by Eq. \eqref{eq_condensed_y}. With the structure and features of the condensed graph fixed, the condensed graph by PreGC shows similar patterns of change as the original graph as the training ratio increases. This label allocation strategy significantly improves the reusability of PreGC.

\begin{wraptable}{r}{7.8cm}\scriptsize
\vspace{-2.em}
\caption{Experimental results for the ablation study.}
\vspace{1.0em}
  \centering
  \begin{tabular}{cccccc}
    \toprule
    \multirow{2}[4]{*}{H\&M} & \multicolumn{2}{c}{$r=2.5\%$} &       & \multicolumn{2}{c}{$r=5.0\%$} \\
    \cmidrule{2-3}\cmidrule{5-6}          & Category & Price &       & Category & Price \\
    \cmidrule{1-4}\cmidrule{5-6}   
    PreGC   & 69.1±0.3 & 34.4±0.9 &       &  71.1±0.3& 36.9±1.1 \\
    PreGC$_{\text{w/o}~Aug}$ & 66.6±0.7 & 32.2±0.9 &       & 67.9±0.4 & 33.3±1.4 \\
    PreGC$_{\text{w/o}~\mathcal{L}_{cost}}$ & 67.5±0.5 & 29.1±0.8 &       & 68.9±0.8 & 31.7±1.5 \\
    \bottomrule
  \end{tabular}
  \label{tab:ablation}
  \vspace{-1.0em}
\end{wraptable}
\textbf{Ablation Study.} To validate the efficacy of crucial optimization strategies, we conduct a comprehensive analysis on H\&M dataset through ablation studies. The experimental results are shown in Table \ref{tab:ablation}, wherein "PreGC$_{\text{w/o}~Aug}$" is the PreGC that removes the graph diffusion augmentation and "PreGC$_{\text{w/o}~\mathcal{L}_{cost}}$" is the adoption of the existing strategy, i.e. predefined node labels before GC. We observe that both graph diffusion augmentation and $\mathcal{L}_{cost}$ contribute significantly to the improvement of the GC optimization. Notably, 
"PreGC$_{\text{w/o}~\mathcal{L}_{cost}}$" induces label dependency in condensation, rendering the condensed graph via category prove inadequate for achieving superior performance on price prediction.
Furthermore, the combination of these two modules increases the stability of the condensed graph.


More experiments and related work are detailed in Appendix D and Appendix E, respectively.


\vspace{-0.05cm}
\section{Conclusions}
\vspace{-0.05cm}
In this work, we revisit the limitations of current GC methods and formulate a generalized GC objective. Following this paradigm, we further propose a pre-trained graph condensation framework for distilling a task-agnostic and architecture-flexible condensed graph.
To avoid excessive dependence of condensed graphs on specific GNNs, a hybrid-interval graph diffusion augmentation is proposed to enhance generalization by increasing node state uncertainty. Subsequently, a plan matching mechanism is derived to eliminate task constraints, which relies on matching optimal transport plans for graph alignment and representation alignment to preserve semantic association consistency. In addition, supervised fine-tuning further refines the condensed task signals and improves the performance of the condensed graph on specific tasks.
The experimental results consistently demonstrate the efficacy and reusability of PreGC, showcasing its generalization on various GNNs and tasks. 





\bibliographystyle{plain}
\bibliography{reference-pgc}

\begin{thebibliography}{10}

\bibitem{DBLP:journals/corr/abs-2409-14500}
Gleb Bazhenov, Oleg Platonov, and Liudmila Prokhorenkova.
\newblock Tabgraphs: {A} benchmark and strong baselines for learning on graphs with tabular node features.
\newblock {\em arXiv preprint arXiv: 2409.14500}, 2024.

\bibitem{DBLP:conf/emnlp/ByunCKP24}
Ju{-}Seung Byun, Jiyun Chun, Jihyung Kil, and Andrew Perrault.
\newblock {ARES:} alternating reinforcement learning and supervised fine-tuning for enhanced multi-modal chain-of-thought reasoning through diverse {AI} feedback.
\newblock In {\em Empirical Methods in Natural Language Processing (EMNLP)}, pages 4410--4430, 2024.

\bibitem{DBLP:conf/www/ChenBSXZHHWH24}
Hao Chen, Yuanchen Bei, Qijie Shen, Yue Xu, Sheng Zhou, Wenbing Huang, Feiran Huang, Senzhang Wang, and Xiao Huang.
\newblock Macro graph neural networks for online billion-scale recommender systems.
\newblock In {\em the {ACM} on Web Conference (WWW)}, pages 3598--3608, 2024.

\bibitem{DBLP:conf/icml/ChenG0LC020}
Liqun Chen, Zhe Gan, Yu~Cheng, Linjie Li, Lawrence Carin, and Jingjing Liu.
\newblock Graph optimal transport for cross-domain alignment.
\newblock In {\em International Conference on Machine Learning (ICML)}, volume 119, pages 1542--1553, 2020.

\bibitem{gprgnn_iclr21}
Eli Chien, Jianhao Peng, Pan Li, and Olgica Milenkovic.
\newblock Adaptive universal generalized pagerank graph neural network.
\newblock In {\em International Conference on Learning Representations (ICLR)}, 2021.

\bibitem{chung1997spectral}
Fan~RK Chung.
\newblock {\em Spectral graph theory}, volume~92.
\newblock American Mathematical Soc., 1997.

\bibitem{DBLP:conf/icde/CuiCYDF024}
Jinhao Cui, Heyan Chai, Xu~Yang, Ye~Ding, Binxing Fang, and Qing Liao.
\newblock {SGCL:} semantic-aware graph contrastive learning with lipschitz graph augmentation.
\newblock In {\em {IEEE} International Conference on Data Engineering (ICDE)}, pages 3028--3041, 2024.

\bibitem{DBLP:conf/icml/FeyHHLR0YYL24}
Matthias Fey, Weihua Hu, Kexin Huang, Jan~Eric Lenssen, Rishabh Ranjan, Joshua Robinson, Rex Ying, Jiaxuan You, and Jure Leskovec.
\newblock Position: Relational deep learning - graph representation learning on relational databases.
\newblock In {\em International Conference on Machine Learning (ICML)}, 2024.

\bibitem{seflgc_arxiv24}
Xinyi Gao, Yayong Li, Tong Chen, Guanhua Ye, Wentao Zhang, and Hongzhi Yin.
\newblock Contrastive graph condensation: Advancing data versatility through self-supervised learning.
\newblock {\em arXiv preprint arXiv: 2411.17063}, 2024.

\bibitem{cgc_www25}
Xinyi Gao, Guanhua Ye, Tong Chen, Wentao Zhang, Junliang Yu, and Hongzhi Yin.
\newblock Rethinking and accelerating graph condensation: A training-free approach with class partition.
\newblock In {\em The ACM Web Conference (WWW)}, 2025.

\bibitem{graphsage_nips17}
William~L. Hamilton, Zhitao Ying, and Jure Leskovec.
\newblock Inductive representation learning on large graphs.
\newblock In {\em Neural Information Processing Systems (NeurIPS)}, pages 1024--1034, 2017.

\bibitem{DBLP:conf/iclr/HanLMT0TY24}
Haoyu Han, Xiaorui Liu, Li~Ma, MohamadAli Torkamani, Hui Liu, Jiliang Tang, and Makoto Yamada.
\newblock Structural fairness-aware active learning for graph neural networks.
\newblock In {\em International Conference on Learning Representations (ICLR)}, 2024.

\bibitem{DBLP:conf/nips/HanF024}
Xiaoxue Han, Zhuo Feng, and Yue Ning.
\newblock A topology-aware graph coarsening framework for continual graph learning.
\newblock In {\em Neural Information Processing Systems (NeurIPS)}, 2024.

\bibitem{bernnet_nips21}
Mingguo He, Zhewei Wei, Zengfeng Huang, and Hongteng Xu.
\newblock Bernnet: Learning arbitrary graph spectral filters via bernstein approximation.
\newblock In {\em Neural Information Processing Systems (NeurIPS)}, pages 14239--14251, 2021.

\bibitem{doscond_kdd22}
Wei Jin, Xianfeng Tang, Haoming Jiang, Zheng Li, Danqing Zhang, Jiliang Tang, and Bing Yin.
\newblock Condensing graphs via one-step gradient matching.
\newblock In {\em Knowledge Discovery and Data Mining (KDD)}, pages 720--730, 2022.

\bibitem{gcond_iclr22}
Wei Jin, Lingxiao Zhao, Shichang Zhang, Yozen Liu, Jiliang Tang, and Neil Shah.
\newblock Graph condensation for graph neural networks.
\newblock In {\em International Conference on Learning Representations (ICLR)}, 2022.

\bibitem{gcn_iclr17}
Thomas~N. Kipf and Max Welling.
\newblock Semi-supervised classification with graph convolutional networks.
\newblock In {\em International Conference on Learning Representations (ICLR)}, 2017.

\bibitem{appnp_iclr19}
Johannes Klicpera, Aleksandar Bojchevski, and Stephan G{\"{u}}nnemann.
\newblock Predict then propagate: Graph neural networks meet personalized pagerank.
\newblock In {\em International Conference on Learning Representations (ICLR)}, 2019.

\bibitem{DBLP:journals/apin/LiLYZZ23}
Chao Li, Xinming Liu, Yeyu Yan, Zhongying Zhao, and Qingtian Zeng.
\newblock Hetgnn-sf: Self-supervised learning on heterogeneous graph neural network via semantic strength and feature similarity.
\newblock {\em Appl. Intell.}, 53(19):21902--21919, 2023.

\bibitem{openfgl_arxiv24}
Xunkai Li, Yinlin Zhu, Boyang Pang, Guochen Yan, Yeyu Yan, Zening Li, Zhengyu Wu, Wentao Zhang, Rong{-}Hua Li, and Guoren Wang.
\newblock Openfgl: {A} comprehensive benchmarks for federated graph learning.
\newblock {\em arXiv preprint arXiv: 2408.16288}, 2024.

\bibitem{diffusion_aaai24}
Yibo Li, Xiao Wang, Hongrui Liu, and Chuan Shi.
\newblock A generalized neural diffusion framework on graphs.
\newblock In {\em {AAAI} Conference on Artificial Intelligence (AAAI)}, pages 8707--8715, 2024.

\bibitem{gcdm_arxiv22}
Mengyang Liu, Shanchuan Li, Xinshi Chen, and Le~Song.
\newblock Graph condensation via receptive field distribution matching.
\newblock {\em arXiv preprint arXiv:2206.13697}, 2022.

\bibitem{gdem_icml24}
Yang Liu, Deyu Bo, and Chuan Shi.
\newblock Graph distillation with eigenbasis matching.
\newblock In {\em International Conference on Machine Learning (ICML)}, 2024.

\bibitem{kgnn_aaai19}
Christopher Morris, Martin Ritzert, Matthias Fey, William~L. Hamilton, Jan~Eric Lenssen, Gaurav Rattan, and Martin Grohe.
\newblock Weisfeiler and leman go neural: Higher-order graph neural networks.
\newblock In {\em {AAAI} Conference on Artificial Intelligence (AAAI)}, pages 4602--4609, 2019.

\bibitem{WD_access25}
Luiz~Manella Pereira and M.~Hadi Amini.
\newblock A survey on optimal transport for machine learning: Theory and applications.
\newblock {\em {IEEE} Access}, 13:26506--26526, 2025.

\bibitem{gcbench_nips24}
Qingyun Sun, Ziying Chen, Beining Yang, Cheng Ji, Xingcheng Fu, Sheng Zhou, Hao Peng, Jianxin Li, and Philip~S. Yu.
\newblock Gc-bench: An open and unified benchmark for graph condensation.
\newblock In {\em Neural Information Processing Systems (NeurIPS)}, 2024.

\bibitem{FGWD_algorithms20}
Titouan Vayer, Laetitia Chapel, R{\'{e}}mi Flamary, Romain Tavenard, and Nicolas Courty.
\newblock Fused gromov-wasserstein distance for structured objects.
\newblock {\em Algorithms}, 13(9):212, 2020.

\bibitem{gat_iclr18}
Petar Velickovic, Guillem Cucurull, Arantxa Casanova, Adriana Romero, Pietro Li{\`{o}}, and Yoshua Bengio.
\newblock Graph attention networks.
\newblock In {\em International Conference on Learning Representations (ICLR)}, 2018.

\bibitem{selfgc_kdd24}
Yuxiang Wang, Xiao Yan, Shiyu Jin, Hao Huang, Quanqing Xu, Qingchen Zhang, Bo~Du, and Jiawei Jiang.
\newblock Self-supervised learning for graph dataset condensation.
\newblock In {\em Knowledge Discovery and Data Mining (KDD)}, pages 3289--3298, 2024.

\bibitem{sgc_icml19}
Felix Wu, Amauri H.~Souza Jr., Tianyi Zhang, Christopher Fifty, Tao Yu, and Kilian~Q. Weinberger.
\newblock Simplifying graph convolutional networks.
\newblock In {\em International Conference on Machine Learning (ICML)}, volume~97, pages 6861--6871, 2019.

\bibitem{simgc_pkdd24}
Zhenbang Xiao, Yu~Wang, Shunyu Liu, Huiqiong Wang, Mingli Song, and Tongya Zheng.
\newblock Simple graph condensation.
\newblock In {\em Machine Learning and Knowledge Discovery in Databases (ECML PKDD)}, volume 14942, pages 53--71, 2024.

\bibitem{DBLP:journals/eswa/YanLYLZ23}
Yeyu Yan, Chao Li, Yanwei Yu, Xiangju Li, and Zhongying Zhao.
\newblock {OSGNN:} original graph and subgraph aggregated graph neural network.
\newblock {\em Expert Syst. Appl.}, 225:120115, 2023.

\bibitem{DBLP:journals/tbd/YanZYYL24}
Yeyu Yan, Zhongying Zhao, Zhan Yang, Yanwei Yu, and Chao Li.
\newblock A fast and robust attention-free heterogeneous graph convolutional network.
\newblock {\em {IEEE} Trans. Big Data}, 10(5):669--681, 2024.

\bibitem{sgdd_nips23}
Beining Yang, Kai Wang, Qingyun Sun, Cheng Ji, Xingcheng Fu, Hao Tang, Yang You, and Jianxin Li.
\newblock Does graph distillation see like vision dataset counterpart?
\newblock In {\em Neural Information Processing Systems (NeurIPS)}, 2023.

\bibitem{DBLP:conf/aaai/Yin024}
Shuo Yin and Guoqiang Zhong.
\newblock Textgt: {A} double-view graph transformer on text for aspect-based sentiment analysis.
\newblock In {\em {AAAI} Conference on Artificial Intelligence (AAAI)}, pages 19404--19412, 2024.

\bibitem{DBLP:conf/kdd/ZhangYS0OLT0022}
Wentao Zhang, Ziqi Yin, Zeang Sheng, Yang Li, Wen Ouyang, Xiaosen Li, Yangyu Tao, Zhi Yang, and Bin Cui.
\newblock Graph attention multi-layer perceptron.
\newblock In {\em Knowledge Discovery and Data Mining (KDD)}, pages 4560--4570, 2022.

\bibitem{geon_icml24}
Yuchen Zhang, Tianle Zhang, Kai Wang, Ziyao Guo, Yuxuan Liang, Xavier Bresson, Wei Jin, and Yang You.
\newblock Navigating complexity: Toward lossless graph condensation via expanding window matching.
\newblock In {\em International Conference on Machine Learning (ICML)}, 2024.

\bibitem{DBLP:conf/kdd/Zhang0Z0Z0025}
Zhongjian Zhang, Xiao Wang, Huichi Zhou, Yue Yu, Mengmei Zhang, Cheng Yang, and Chuan Shi.
\newblock Can large language models improve the adversarial robustness of graph neural networks?
\newblock In {\em Knowledge Discovery and Data Mining (KDD)}, pages 2008--2019, 2025.

\bibitem{DBLP:journals/pami/ZhengZLLZ24}
Shuai Zheng, Zhenfeng Zhu, Zhizhe Liu, Youru Li, and Yao Zhao.
\newblock Node-oriented spectral filtering for graph neural networks.
\newblock {\em {IEEE} Trans. Pattern Anal. Mach. Intell.}, 46(1):388--402, 2024.

\bibitem{sfgc_nips23}
Xin Zheng, Miao Zhang, Chunyang Chen, Quoc Viet~Hung Nguyen, Xingquan Zhu, and Shirui Pan.
\newblock Structure-free graph condensation: From large-scale graphs to condensed graph-free data.
\newblock In {\em Neural Information Processing Systems (NeurIPS)}, 2023.

\bibitem{ssgc_iclr21}
Hao Zhu and Piotr Koniusz.
\newblock Simple spectral graph convolution.
\newblock In {\em International Conference on Learning Representations (ICLR)}, 2021.

\end{thebibliography}

\end{document}